\documentclass{article}
\usepackage{spconf,amsmath,graphicx,hyperref}

\usepackage{enumitem}
\usepackage{cite}
\usepackage{amsmath,amssymb,amsfonts}
\usepackage{algorithmic}
\usepackage{graphicx}
\usepackage{textcomp}
\usepackage{amsmath,graphicx}
\usepackage{booktabs}

\usepackage{setspace}

\usepackage{multirow}
\usepackage{xcolor}
\def\BibTeX{{\rm B\kern-.05em{\sc i\kern-.025em b}\kern-.08em
    T\kern-.1667em\lower.7ex\hbox{E}\kern-.125emX}}

\usepackage{fontawesome5}
\usepackage[capitalise]{cleveref}

\usepackage{xspace}
\usepackage{xcolor}
\usepackage[normalem]{ulem}
\useunder{\uline}{\ulined}{}%
\DeclareUrlCommand{\bulurl}{}
\usepackage[table,dvipsnames]{xcolor}

\begin{document}

\title{AuditoryBench++: Can Language Models Understand Auditory Knowledge without Hearing?}

\name{Hyunjong Ok$^{1,2\ast}$\quad Suho Yoo$^{2,3\ast}$\quad Hyeonjun Kim$^{1}$\quad Jaeho Lee$^{1}$}

\address{
$^1$Pohang University of Science and Technology, South Korea $^2$HJ AILAB \\
$^3$Korea Advanced Institute of Science and Technology, South Korea \\
\thanks{
\begin{tabular}{@{}l}
$^\ast$Equal contribution \\
Correspondence to: Jaeho Lee $<$jaeho.lee@postech.ac.kr$>$.
\end{tabular}
}
}
\maketitle

\begin{abstract}    
Even without directly hearing sounds, humans can effortlessly reason about auditory properties, such as pitch, loudness, or sound-source associations, drawing on auditory commonsense.
In contrast, language models often lack this capability, limiting their effectiveness in multimodal interactions.
As an initial step to address this gap, we present AuditoryBench++, a comprehensive benchmark for evaluating auditory knowledge and reasoning in text-only settings. The benchmark encompasses tasks that range from basic auditory comparisons to contextually grounded reasoning, enabling fine-grained analysis of how models process and integrate auditory concepts. 
In addition, we introduce AIR-CoT, a novel auditory imagination reasoning method that generates and integrates auditory information during inference through span detection with special tokens and knowledge injection.
Extensive experiments with recent LLMs and Multimodal LLMs demonstrate that AIR-CoT generally outperforms both the off-the-shelf models and those augmented with auditory knowledge.
The project page is available at \bulurl{https://auditorybenchpp.github.io}.
\end{abstract}


\section{Introduction}
\label{sec:intro}
Imagination is considered a ``necessary ingredient of perception itself'' in Kantian perspectives \cite{Strawson1974, Gregory2010}, as it allows us to reconstruct a complete representation of the target object out of incomplete, raw sensations. For example, by reading a text describing a night with heavy rain and lightning, one can synthesize multimodal imagery of the scene by imagining the sounds of rain pounding like drumbeats and roaring thunder. Given the same text, people can imagine similar multimodal signals as they share similar experiences and understandings about objects---\textit{i.e.,} a \textit{commonsense}. Having such common sense enables efficient and effective communication between people, without having to describe every detail.

Do large language models (LLMs) share a similar commonsense? Recent studies reveal that the answer is no; LLMs have a poor understanding of both visual commonsense (\textit{e.g.}, colors of everyday objects) \cite{zhang2022visual, liu2022things, alper2023bert} and auditory commonsense (\textit{e.g.}, animal-sound associations) \cite{ok2025audiobert}.
While visual knowledge in LLMs has received significant research focus~\cite{tan2020vokenization, wangvisually, liumind, wu2024mind, liimagine}, auditory knowledge remains critically underexplored, with existing benchmarks and methods being notably scarce. While recent advances in large audio-language models (LALMs) have shown promising results when processing audio inputs, their ability to engage in such auditory imagination in purely text-only settings---where no audio signal is available---remains underexplored. We need to start with tasks that are trivial for humans yet require basic auditory imagination, providing an initial basis for systematically assessing such abilities in LLMs as well as LALMs.

\begin{figure}[t]
    \centering
    \includegraphics[width=1\linewidth]{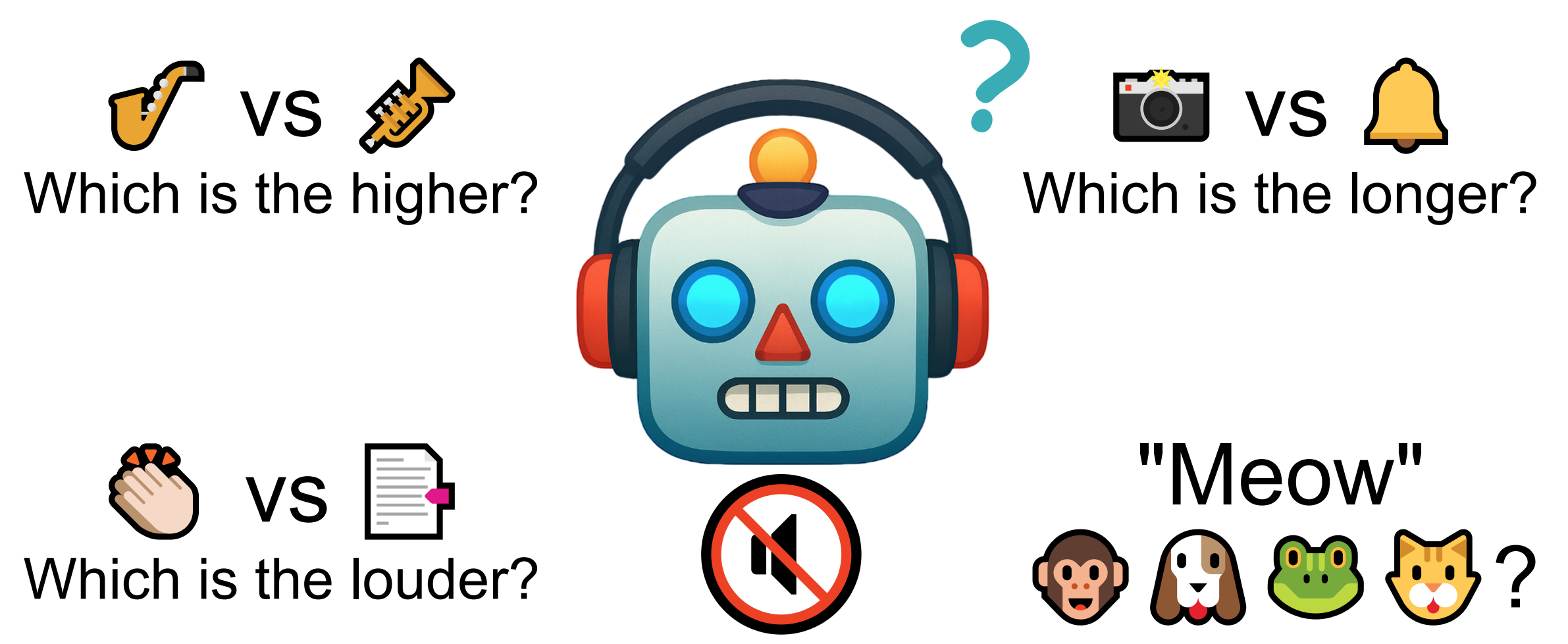}
    \caption{Overview of \textbf{AuditoryBench++},
which assesses auditory knowledge of language models without audio input.}
    \label{fig:fig1}
\end{figure}

For this purpose, we introduce AuditoryBench++, a comprehensive benchmark for evaluating auditory knowledge in LLMs. Unlike previous benchmarks~\cite{ok2025audiobert} limited to short-answer simple inference tasks across only two categories, our benchmark encompasses five tasks, including reasoning tasks, providing a more thorough assessment of auditory reasoning capabilities. We refine and verify the dataset through rigorous processes, ensuring superior quality and reliability over previous benchmarks. Through this benchmark, we figure out that both current LLMs and LALMs (w/o audio input) perform close to random guessing in comparison tasks.

We propose a novel auditory reasoning method that enables LLMs to dynamically generate and process auditory embedding during inference, enabling LLMs to seemingly hear through explicit auditory imagination. We introduce an innovative training paradigm that equips LLMs with the ability to \textit{imagine} sounds when encountering auditory reasoning tasks. Specifically, we introduce a special token, {\footnotesize\tt[imagine]}\xspace, emitted by the model whenever auditory reasoning is required. When the model encounters contexts that require auditory reasoning, it learns to emit this token, signaling the system to pause text generation and invoke an imagination process to think contextually relevant sounds. The model then integrates this auditory information to continue reasoning with enhanced acoustic understanding. Unlike the cascade approach in previous methods such as Imagine to Hear~\cite{yoo2025imagine}, our end-to-end method empowers the model to reason, achieving superior performance.



\section{Related work}

\paragraph*{Multimodal commonsense benchmark.} 

Multimodal commonsense benchmarks assess whether models can combine perceptual knowledge with language understanding. In the vision domain, various datasets evaluate visual commonsense, such as knowledge grounding or compositional reasoning~\cite{marino2019ok, shen-etal-2025-vcd, yue2024mmmu}. For the audio modality, there are large-scale resources for sound event classification, captioning, and reasoning~\cite{xie2025audiotime, huh2025epic, gemmeke2017audio, sakshi2025mmau}. Taking this further, several studies have extended this evaluation to audio-visual tasks, requiring reasoning over audio and visual cues~\cite{yang2022avqa, gong2024av}. 

However, these benchmarks generally assume perceptual inputs (image, audio, or both). In contrast, AuditoryBench++ targets text-only settings, requiring models to imagine auditory properties and isolating their internal commonsense without relying on perceptual inputs to provide cues. This design probes whether models can still reason about sounds when direct audio input is unavailable, reflecting real-world cases where only textual descriptions are accessible.

\begin{figure*}[t]
    \centering
    \includegraphics[width=1\linewidth]{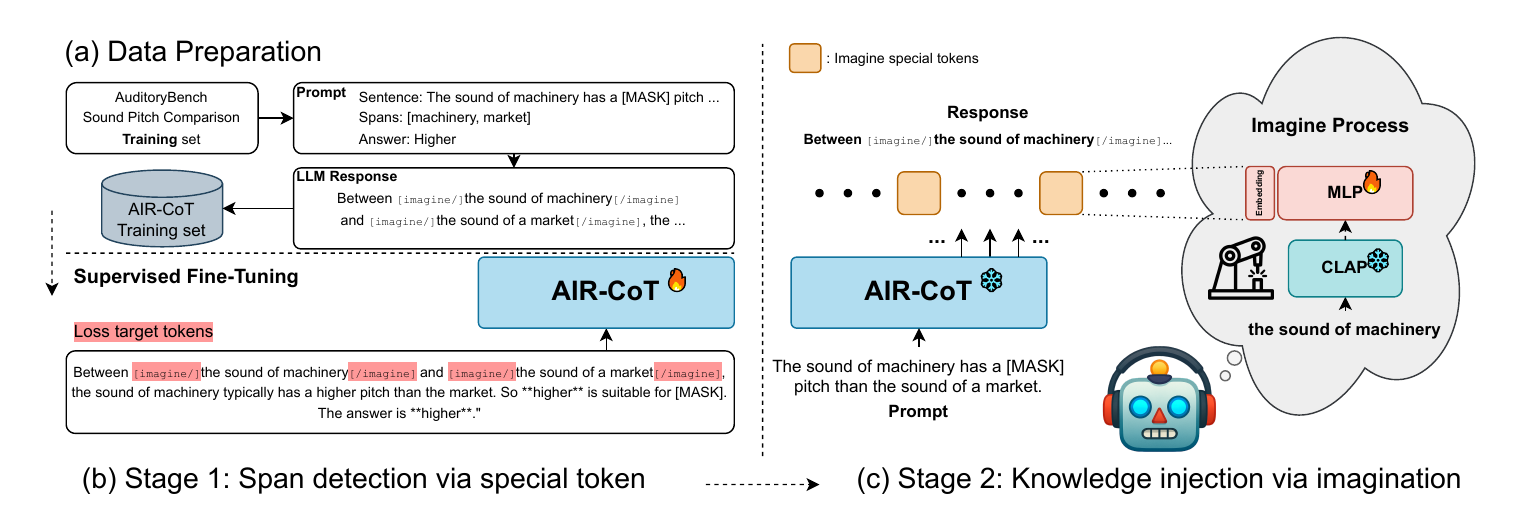}
    \caption{Pipeline of the proposed \textbf{AIR-CoT}. 
(a) Data Preparation. Training data is augmented with {\footnotesize\tt[imagine]}\xspace tokens to mark spans requiring auditory reasoning.  
(b) Stage 1: Span Detection. The model is fine-tuned to detect the spans by generating the special tokens during decoding.  
(c) Stage 2: Knowledge Injection. When encountering the {\footnotesize\tt[/imagine]}\xspace token, the model pauses to generate the embedding using CLAP and injects it for auditory reasoning.
} 
\label{fig:method}
\end{figure*}

\paragraph*{Reasoning models via imagination.}

Reasoning models enhance the inference capabilities of LLMs, thereby expanding their overall capacity~\cite{guo2025deepseek, liu2025visual}. While reinforcement learning tuning remains the dominant approach for enabling reasoning, recent methods have explored imagination-based alternatives~\cite{liumind, zhou2024minedreamer}.
These methods involve visual thought processes, such as generating subsequent scenes or editing given images~\cite{wu2024mind, liimagine, xu2025visual}, which leverage visual imagination as an intermediate reasoning step---demonstrating improved performance on multi-step reasoning tasks. 

In contrast to existing visual imagination frameworks, our approach explores auditory imagination as a novel modality for enhancing reasoning capabilities, introducing auditory thoughts as a complementary tool for complex inference.

\section{AuditoryBench++}

\paragraph*{Task Definition.}
AuditoryBench++ comprises 5 tasks evaluating a spectrum of auditory knowledge, from fundamental comparisons to complex, contextually grounded reasoning:

\begin{itemize}[leftmargin=*,topsep=0pt,parsep=0.2pt]

    \item \textit{Pitch Comparison}:  
    The model selects which of two sounds has a higher pitch, formulated as a binary decision task.

    \item \textit{Duration Comparison}:  
    The model compares two described sounds and identifies the one with the longer duration.

    \item \textit{Loudness Comparison}:  
    This task asks the model to select the louder sound between two options based on prompts.

    \item \textit{Animal Sound Recognition}:  
    This task requires predicting the correct animal corresponding to a given onomatopoeic expression (e.g., 'meow'). Each sample is presented as a multiple-choice question with four options.

    \item \textit{Auditory Context Reasoning}: 
    This component evaluates a model's ability to perform contextual auditory reasoning, focusing on interpreting nuanced auditory cues and situational contexts in a multiple-choice format.
    
\end{itemize}


\begin{table}[t]
\centering
\caption{Task summary of \textbf{AuditoryBench++} reconstructed from original resources with filtering and verification.}
\label{tab:auditorybench_tasks}
\resizebox{1\columnwidth}{!}{%
\begin{tabular}{@{} l l l l r @{}}
\toprule
\textbf{Task Type} & \textbf{Task} & \textbf{Original Resource} & \textbf{Question Type} & \textbf{\# QnA} \\
\midrule
Comparison & Pitch & AuditoryBench & Binary Choice & 3,625 \\
           & Duration & AudioTime &  & 1,645 \\
           & Loudness & AudioTime &  & 445 \\
\midrule
Recognition & Animal Sound & AuditoryBench & Multiple Choice & 942 \\
\midrule
Reasoning   & Auditory Context & MMAU & Multiple Choice & 75 \\
\midrule
\multicolumn{4}{@{}l}{\textbf{Total}} & \textbf{6,732} \\
\bottomrule
\end{tabular}%
}
\end{table}



 \paragraph*{Construction Pipeline.}
To construct AuditoryBench++, we carefully design a multi-stage pipeline that integrates diverse existing resources and applies systematic filtering, statistical estimation, and human verification. This process ensures the resulting tasks are both reliable and unambiguous, providing a robust benchmark for auditory reasoning.


For pitch comparison, we use only the wiki set of AuditoryBench~\cite{ok2025audiobert}, since it consists of instrument-based pitch pairs that ensure objectivity and consistency in pairwise evaluation. For animal sound recognition, we draw from both the test and wiki sets of AuditoryBench, and apply human filtering to both choices and answers. Inappropriate or ambiguous samples (e.g.,  fictional animals or mislabeled cases) are removed to ensure the reliability of the evaluation.

For duration and loudness comparison, we build new datasets from AudioTime~\cite{xie2025audiotime}, leveraging its segment-level annotations. In the loudness task, peak decibel levels are calculated to provide a consistent measure of intensity across samples, ensuring clear distinctions between label pairs and minimizing ambiguity. We first retain only data classes with at least 30 samples to ensure statistical reliability. Outliers are removed using the interquartile range (IQR) rule, and we compute pairwise differences to select label pairs with statistically significant contrasts ($p < 0.01$). The final dataset consists of comparison questions between label pairs that exhibit meaningful and measurable differences.

To build the auditory context reasoning task, we start from the open set of MMAU~\cite{sakshi2025mmau} and adapt its audio-related reasoning questions into a text-only format. Each audio clip is first described using Qwen2-Audio~\cite{chu2024qwen2}, generating detailed captions that capture the key auditory cues, such as sound sources, events, and acoustic properties. These captions are then paired with the original questions and rewritten by GPT-4o~\cite{hurst2024gpt} to create text-only problems that retain the original reasoning objectives. Subsequently, human verification carefully removes unnatural or erroneous rewrites and revises them to ensure the questions are precise, coherent, and natural in a purely text-based setting.

At the end of each dataset construction pipeline, all task datasets underwent a final round of human filtering and refinement to ensure correctness and remove residual noise.

\paragraph*{Omitted details.}
Certain details (\textit{e.g.}, details of the pipeline, dataset samples, etc.) are omitted here for brevity. 
These can be found on our project page\footnote{\url{https://auditorybenchpp.github.io}}.

\section{AIR-CoT}
In this section, we introduce the auditory imagination reasoning Chain-of-Thought (AIR-CoT), a novel method to equip language models with auditory capabilities and thereby enable reasoning grounded in auditory commonsense.

AIR-CoT proceeds in two training stages. In the first stage, we apply SFT to train the model to detect spans requiring auditory knowledge during decoding via special tokens. In the second stage, we train an imagination module to inject relevant auditory knowledge into these identified spans, enabling the model to perform auditory reasoning.

\paragraph*{Data preparation.}
Before training, we curate data to train our model. Drawing from the training set of the sound pitch comparison task in AuditoryBench~\cite{ok2025audiobert},
which is independent from our pitch comparison data with no overlapping sound objects,
we leverage an LLM to generate reasoning-style outputs. As shown in Fig. 2(a), these outputs encompass spans that require auditory knowledge within {\footnotesize\tt[imagine]}\xspace tokens. Specifically, we employ the Qwen2.5-32B~\cite{qwen2025qwen25technicalreport} model for this generation process, providing it with few-shot examples and the current data's information; context, span, and answer.

\paragraph*{Stage 1: Span detection via special token.}
To enable models to imagine and reason with auditory knowledge, we first train them to detect spans that require such knowledge. We introduce special {\footnotesize\tt[imagine]}\xspace tokens, which the model generates to detect auditory knowledge demanded spans during reasoning (\textit{e.g.}, {\footnotesize\tt[imagine/]}\xspace the sound of machinery {\footnotesize\tt[/imagine]}\xspace). We perform SFT with the loss focusing solely on generating these tokens. Downstream answer tokens (\textit{e.g.}, `higher' in a pitch comparison task) are excluded from the loss to prioritize accurate span detection without biasing final predictions.

\begin{table*}[!]
\caption{Experimental results of AIR-CoT across comparison, recognition, and reasoning tasks.}
\centering
\vspace{0.20cm}
\resizebox{0.75\textwidth}{!}{%
\begin{tabular}{l|ccc|c|c}
\toprule
\multirow{2}{*}{\textbf{Methods}} & 
\multicolumn{3}{c|}{\textbf{Comparison}} & 
\multicolumn{1}{c|}{\textbf{Recognition}} & 
\multicolumn{1}{c}{\textbf{Reasoning}} \\ 
& \multicolumn{1}{c}{Pitch} & Duration & Loudness & \multicolumn{1}{c}{Animal Sound} & \multicolumn{1}{c}{Auditory Context} \\ 
\midrule
Majority Class & 52.14 & 56.11 & 57.30 & 26.65 & 25.33 \\
\midrule 
\multicolumn{6}{c}{Off-the-Shelf LLMs}\\
\midrule
LLaMA3.1$_{8\mathrm{B}}$ \cite{dubey2024llama} & 52.39 & 55.81 & 46.29 & 56.26 & 69.33 \\
Qwen2.5$_{7\mathrm{B}}$ \cite{qwen2025qwen25technicalreport} & 42.46 & 55.44 & 57.53 & 62.21 & 70.67 \\
Phi4-mini$_{4\mathrm{B}}$ \cite{abdin2024phi} & 55.75 & 54.59 & 58.65 & 60.72 & 68.00 \\
Qwen3-30B$_{30\mathrm{B}}$ \cite{yang2025qwen3} & 57.21 & 53.68 & 50.11 & 72.40 & 72.00 \\
Qwen3-Omni$_{30\mathrm{B}}$ \cite{Qwen3-Omni} & 62.37 & 54.29 & 53.48 & \textbf{73.14} & 73.33 \\
Qwen2-Audio$_{7\mathrm{B}}$ \cite{chu2024qwen2} & 46.59 & 50.40 & 55.73 & 31.42 & 37.33 \\
Phi4-MM$_{6\mathrm{B}}$ \cite{abouelenin2025phi} & 56.08 & 56.84 & 57.98 & 60.93 & 65.33 \\
\midrule 
\multicolumn{6}{c}{Augmented Methods}\\
\midrule
AudioBERT \cite{ok2025audiobert} & 59.34 & 51.91 & 57.30 & 22.40 & 30.67 \\
Imagine to Hear \cite{yoo2025imagine} & 75.64 & \textbf{58.78} & 57.30 & 24.73 & 29.33 \\
Standard SFT (Qwen2.5$_{7\mathrm{B}}$) & 75.28 & 42.07 & 55.96 & 57.86 & 61.33 \\
\midrule
AIR-CoT (Ours) & 
\textbf{83.89} {\color{ForestGreen}(+8.25)} & 
54.59 {\color{red}(-4.19)} & 
\textbf{59.33} {\color{ForestGreen}(+0.68)} & 
71.55 {\color{red}(-1.59)} & 
\textbf{82.67} {\color{ForestGreen}(+9.34)} \\
\bottomrule
\end{tabular}
}
\label{tab:main}
\end{table*}

\paragraph*{Stage 2: Knowledge injection via imagination.}
After span detection, we equip the model with the ability to imagine auditory knowledge by injecting suitable embeddings. During decoding, upon generating the {\footnotesize\tt[/imagine]} token, the model pauses to perform auditory imagination. We leverage audio models (e.g., CLAP~\cite{elizalde2023clap}) to produce audio embeddings and inject them into the {\footnotesize\tt[/imagine]} token. This enables AIR-CoT; when the model needs auditory knowledge, it stops, imagines it, and then continues the CoT.

In detail, we adapt the embeddings via a 2-layer MLP to match the model's hidden size. Training focuses solely on the MLP, with loss computed only on answer tokens.

\section{Experiments}
\paragraph*{Baselines and metric.}

We establish comprehensive baselines for AuditoryBench++, covering recent LLMs, audio-language models (LALMs), and prior methods that augment or inject auditory knowledge. 
For LLMs, we evaluate LLaMA3.1$_{8\mathrm{B}}$ \cite{dubey2024llama}, Qwen2.5$_{7\mathrm{B}}$ \cite{qwen2025qwen25technicalreport}, Phi4-mini$_{4\mathrm{B}}$ \cite{abdin2024phi}, and Qwen3-30B$_{30\mathrm{B}}$ \cite{yang2025qwen3}. 
For LALMs, we include Qwen2-Audio$_{7\mathrm{B}}$ \cite{chu2024qwen2}, Phi4-MM$_{6\mathrm{B}}$ \cite{abouelenin2025phi}, and Qwen3-Omni$_{30\mathrm{B}}$ \cite{Qwen3-Omni}. 
Additionally, we compare our approach with auditory knowledge injection methods such as AudioBERT~\cite{ok2025audiobert} and Imagine to Hear~\cite{yoo2025imagine}. Also report results from a standard supervised fine-tuning baseline on Qwen2.5$_{7\mathrm{B}}$). 
All evaluations utilize accuracy as the metric.

\paragraph*{Implementation details.} 
In our implementation, we base AIR-CoT on the Qwen2.5$_{7\mathrm{B}}$ model for both stages. The imagination process is built with a CLAP text encoder and a 2-layer MLP. 
For Stage 1, we perform fine-tuning with special tokens {\footnotesize\tt[imagine/]}\xspace and {\footnotesize\tt[/imagine]}\xspace added to the vocabulary. Training uses with \(10\) epochs, batch size of \(4\), gradient accumulation steps of \(16\), learning rate of \(1 \times 10^{-5}\), and the AdamW~\cite{loshchilov2018decoupled} optimizer. 
For Stage 2, we load the fine-tuned model from Stage 1 and integrate a CLAP encoder with a 2-layer MLP projector to align audio embeddings. We train only the projector for \(10\) epochs, with a batch size of \(4\), a learning rate of \(1 \times 10^{-4}\), a weight decay of \(0.01\), and an AdamW optimizer. For fair comparison, AudioBERT~\cite{ok2025audiobert} and Imagine to Hear~\cite{yoo2025imagine} employ the same data.

\paragraph*{Experimental results.} 
We report the performance of various baselines and our method in AuditoryBench++.
As shown in~\Cref{tab:main}, most language models perform poorly due to the absence of auditory knowledge. AudioBERT and Imagine to Hear provide some improvements but remain constrained. In contrast, AIR-CoT substantially outperforms these on pitch comparison, animal sound recognition, and auditory context reasoning, showcasing its ability to enable end-to-end auditory imagination within the reasoning chain. 
However, the improvements in duration and loudness remain limited. Current audio representations are primarily semantic, which makes them effective for animal sound recognition and auditory context reasoning. Pitch benefits from exposure to pitch-related training data, enabling relation reasoning.
In contrast, duration and loudness depend on temporal and amplitude cues that these representations do not capture well, with duration being particularly challenging since current embeddings do not reflect the time axis~\cite{ ma2023crepevisionlanguagefoundationmodels, sinha2021maskedlanguagemodelingdistributional}.
We leave this challenging problem for future work.


\section{Conclusions}
We presented AuditoryBench++, a benchmark for assessing auditory knowledge in text-only settings, and AIR-CoT, a reasoning method that equips LLMs with auditory imagination via span detection and knowledge injection. Experiments demonstrate that AIR-CoT outperforms off-the-shelf and augmented models. We believe our work provides a strong foundation for building language models that can imagine auditory information without direct audio input, ultimately enabling more natural and human-like multimodal reasoning.



\clearpage

\section*{Acknowledgements}
This work was supported by the Institute of Information \& Communications Technology Planning \& Evaluation (IITP) grant funded by the Korea government (MSIT) (No.~RS-2024-00457882, AI Research Hub Project), and by the National Research Foundation of Korea (NRF) grant funded by the Korea government (MSIT) (No.~RS-2024-00453301).

\apptocmd{\thebibliography}{%
  \setlength{\itemsep}{0pt}
  \setlength{\parskip}{0pt}
}{}{}

\begingroup
\setstretch{0.75} 
\bibliographystyle{IEEEtran}
\bibliography{shortstrings,refs}

@inproceedings{ok2025audiobert,
  title={AudioBERT: Audio knowledge augmented language model},
  author={Ok, Hyunjong and Yoo, Suho and Lee, Jaeho},
  booktitle=icassp,
  year={2025},
}

@inproceedings{xie2025audiotime,
  title={Audiotime: A temporally-aligned audio-text benchmark dataset},
  author={Xie, Zeyu and Xu, Xuenan and Wu, Zhizheng and Wu, Mengyue},
  booktitle=icassp,
  year={2025},
}

@article{huh2025epic,
  title={Epic-sounds: A large-scale dataset of actions that sound},
  author={Huh, Jaesung and Chalk, Jacob and Kazakos, Evangelos and Damen, Dima and Zisserman, Andrew},
  journal=tpami,
  year={2025},
}

@article{Gregory2010,
  author  = {Gregory, D.},
  title   = {Imagery, the Imagination and Experience},
  journal = {Philosophical Quarterly},
  year    = {2010}
}

@incollection{Strawson1974,
  author    = {Strawson, P. F.},
  title     = {Imagination and Perception},
  booktitle = {Freedom and Resentment},
  publisher = {Methuen},
  address   = {London},
  year      = {1974}
}

@inproceedings{liumind,
  title={Mind's Eye: Grounded Language Model Reasoning through Simulation},
  author={Liu, Ruibo and Wei, Jason and Gu, Shixiang Shane and Wu, Te-Yen and Vosoughi, Soroush and Cui, Claire and Zhou, Denny and Dai, Andrew M},
  booktitle=iclr,
  year={2023},
}

@article{zhou2024minedreamer,
  title={Minedreamer: Learning to follow instructions via chain-of-imagination for simulated-world control},
  author={Zhou, Enshen and Qin, Yiran and Yin, Zhenfei and Huang, Yuzhou and Zhang, Ruimao and Sheng, Lu and Qiao, Yu and Shao, Jing},
  journal={arXiv preprint arXiv:2403.12037},
  year={2024},
}

@inproceedings{wu2024mind,
  title={Mind's eye of LLMs: visualization-of-thought elicits spatial reasoning in large language models},
  author={Wu, Wenshan and Mao, Shaoguang and Zhang, Yadong and Xia, Yan and Dong, Li and Cui, Lei and Wei, Furu},
  booktitle=neurips,
  year={2024},
}

@inproceedings{liimagine,
  title={Imagine While Reasoning in Space: Multimodal Visualization-of-Thought},
  author={Li, Chengzu and Wu, Wenshan and Zhang, Huanyu and Xia, Yan and Mao, Shaoguang and Dong, Li and Vuli{\'c}, Ivan and Wei, Furu},
  booktitle=icml,
  year={2025},
}

@inproceedings{xu2025visual,
  title={Visual Planning: Let's Think Only with Images},
  author={Xu, Yi and Li, Chengzu and Zhou, Han and Wan, Xingchen and Zhang, Caiqi and Korhonen, Anna and Vuli{\'c}, Ivan},
  booktitle=cvprw,
  year={2025},
}

@article{guo2025deepseek,
  title={Deepseek-r1: Incentivizing reasoning capability in llms via reinforcement learning},
  author={Guo, Daya and Yang, Dejian and Zhang, Haowei and Song, Junxiao and Zhang, Ruoyu and Xu, Runxin and Zhu, Qihao and Ma, Shirong and Wang, Peiyi and Bi, Xiao and others},
  journal={arXiv preprint arXiv:2501.12948},
  year={2025},
}

@article{liu2025visual,
  title={Visual-rft: Visual reinforcement fine-tuning},
  author={Liu, Ziyu and Sun, Zeyi and Zang, Yuhang and Dong, Xiaoyi and Cao, Yuhang and Duan, Haodong and Lin, Dahua and Wang, Jiaqi},
  journal={arXiv preprint arXiv:2503.01785},
  year={2025},
}

@inproceedings{marino2019ok,
  title={Ok-vqa: A visual question answering benchmark requiring external knowledge},
  author={Marino, Kenneth and Rastegari, Mohammad and Farhadi, Ali and Mottaghi, Roozbeh},
  booktitle=cvpr,
  year={2019},
}

@inproceedings{shen-etal-2025-vcd,
  title={VCD: A Dataset for Visual Commonsense Discovery in Images},
  author={Shen, Xiangqing and Wang, Fanfan and Wu, Siwei and Xia, Rui},
  booktitle=acl,
  year={2025},
}

@inproceedings{gemmeke2017audio,
  title={Audio set: An ontology and human-labeled dataset for audio events},
  author={Gemmeke, Jort F and Ellis, Daniel PW and Freedman, Dylan and Jansen, Aren and Lawrence, Wade and Moore, R Channing and Plakal, Manoj and Ritter, Marvin},
  booktitle=icassp,
  year={2017},
}

@inproceedings{yang2022avqa,
  title={Avqa: A dataset for audio-visual question answering on videos},
  author={Yang, Pinci and Wang, Xin and Duan, Xuguang and Chen, Hong and Hou, Runze and Jin, Cong and Zhu, Wenwu},
  booktitle=acmmm,
  year={2022},
}

@article{gong2024av,
  title={AV-Odyssey Bench: Can Your Multimodal LLMs Really Understand Audio-Visual Information?},
  author={Gong, Kaixiong and Feng, Kaituo and Li, Bohao and Wang, Yibing and Cheng, Mofan and Yang, Shijia and Han, Jiaming and Wang, Benyou and Bai, Yutong and Yang, Zhuoran and others},
  journal={arXiv preprint arXiv:2412.02611},
  year={2024},
}

@inproceedings{sakshi2025mmau,
  title={{MMAU}: A Massive Multi-Task Audio Understanding and Reasoning Benchmark},
  author={S Sakshi and Utkarsh Tyagi and Sonal Kumar and Ashish Seth and Ramaneswaran Selvakumar and Oriol Nieto and Ramani Duraiswami and Sreyan Ghosh and Dinesh Manocha},
  booktitle=iclr,
  year={2025},
}

@article{abdin2024phi,
  title={Phi-4 technical report},
  author={Abdin, Marah and Aneja, Jyoti and Behl, Harkirat and Bubeck, S{\'e}bastien and Eldan, Ronen and Gunasekar, Suriya and Harrison, Michael and Hewett, Russell J and Javaheripi, Mojan and Kauffmann, Piero and others},
  journal={arXiv preprint arXiv:2412.08905},
  year={2024}
}

@article{dubey2024llama,
  title={The llama 3 herd of models},
  author={Dubey, A. and Jauhri, A. and Pandey, A. and Kadian, A. and Al-Dahle, A. and Letman, A. and Mathur, A. and Schelten, A. and Yang, A. and Fan, A. and others},
  journal={arXiv preprint arXiv:2407.21783},
  year={2024}
}

@article{qwen2025qwen25technicalreport,
  title={Qwen2.5 Technical Report},
  author={Team, Qwen},
  journal={arXiv preprint arXiv:2412.15115},
  year={2024}
}

@article{chu2024qwen2,
  title={Qwen2-audio technical report},
  author={Chu, Yunfei and Xu, Jin and Yang, Qian and Wei, Haojie and Wei, Xipin and Guo, Zhifang and Leng, Yichong and Lv, Yuanjun and He, Jinzheng and Lin, Junyang and others},
  journal={arXiv preprint arXiv:2407.10759},
  year={2024}
}

@article{abouelenin2025phi,
  title={Phi-4-mini technical report: Compact yet powerful multimodal language models via mixture-of-loras},
  author={Abouelenin, Abdelrahman and Ashfaq, Atabak and Atkinson, Adam and Awadalla, Hany and Bach, Nguyen and Bao, Jianmin and Benhaim, Alon and Cai, Martin and Chaudhary, Vishrav and Chen, Congcong and others},
  journal={arXiv preprint arXiv:2503.01743},
  year={2025}
}

@inproceedings{yoo2025imagine,
  title={Imagine to Hear: Auditory Knowledge Generation can be an Effective Assistant for Language Models},
  author={Yoo, Suho and Ok, Hyunjong and Lee, Jaeho},
  booktitle=aclf,
  year={2025},
}

@article{hurst2024gpt,
  title={Gpt-4o system card},
  author={Hurst, Aaron and Lerer, Adam and Goucher, Adam P and Perelman, Adam and Ramesh, Aditya and Clark, Aidan and Ostrow, AJ and Welihinda, Akila and Hayes, Alan and Radford, Alec and others},
  journal={arXiv preprint arXiv:2410.21276},
  year={2024}
}

@inproceedings{yue2024mmmu,
  title={Mmmu: A massive multi-discipline multimodal understanding and reasoning benchmark for expert agi},
  author={Yue, Xiang and Ni, Yuansheng and Zhang, Kai and Zheng, Tianyu and Liu, Ruoqi and Zhang, Ge and Stevens, Samuel and Jiang, Dongfu and Ren, Weiming and Sun, Yuxuan and others},
  booktitle=cvpr,
  year={2024}
}

@inproceedings{zhang2022visual,
  title={Visual Commonsense in Pretrained Unimodal and Multimodal Models},
  author={Zhang, C. and Van Durme, B. and Li, Z. and Stengel-Eskin, E.},
  booktitle=naacl,
  year={2022}
}

@inproceedings{liu2022things,
  title={Things not Written in Text: Exploring Spatial Commonsense from Visual Signals},
  author={Liu, X. and Yin, D. and Feng, Y. and Zhao, D.},
  booktitle=acl,
  year={2022}
}

@inproceedings{alper2023bert,
  title={Is bert blind? exploring the effect of vision-and-language pretraining on visual language understanding},
  author={Alper, M. and Fiman, M. and Averbuch-Elor, H.},
  booktitle=cvpr,
  year={2023}
}

@inproceedings{wangvisually,
  title={Visually-Augmented Language Modeling},
  author={Wang, W. and Dong, L. and Cheng, H. and Song, H. and Liu, X. and Yan, X. and Gao, J. and Wei, F.},
  booktitle=iclr,
  year={2023}
}

@inproceedings{tan2020vokenization,
  title={Vokenization: Improving Language Understanding with Contextualized, Visual-Grounded Supervision},
  author={Tan, H. and Bansal, M.},
  booktitle=emnlp,
  year={2020}
}

@inproceedings{
    loshchilov2018decoupled,
    title={Decoupled Weight Decay Regularization},
    author={I. Loshchilov. and . Hutter},
    booktitle=iclr,
    year={2019},
}

@inproceedings{elizalde2023clap,
  title={Clap learning audio concepts from natural language supervision},
  author={Elizalde, B. and Deshmukh, S. and Al Ismail, M. and Wang, H.},
  booktitle=icassp,
  year={2023}
}

@inproceedings{ma2023crepevisionlanguagefoundationmodels,
      title={CREPE: Can Vision-Language Foundation Models Reason Compositionally?}, 
      author={Zixian Ma and Jerry Hong and Mustafa Omer Gul and Mona Gandhi and Irena Gao and Ranjay Krishna},
      year={2023},
      booktitle=cvpr
}

@inproceedings{sinha2021maskedlanguagemodelingdistributional,
      title={Masked Language Modeling and the Distributional Hypothesis: Order Word Matters Pre-training for Little}, 
      author={Koustuv Sinha and Robin Jia and Dieuwke Hupkes and Joelle Pineau and Adina Williams and Douwe Kiela},
      year={2021},
      booktitle=emnlp
}

@article{yang2025qwen3,
  title={Qwen3 technical report},
  author={Yang, An and Li, Anfeng and Yang, Baosong and Zhang, Beichen and Hui, Binyuan and Zheng, Bo and Yu, Bowen and Gao, Chang and Huang, Chengen and Lv, Chenxu and others},
  journal={arXiv preprint arXiv:2505.09388},
  year={2025}
}

@article{Qwen3-Omni,
  title={Qwen3-Omni Technical Report},
  author={Jin Xu and Zhifang Guo and Hangrui Hu and Yunfei Chu and Xiong Wang and Jinzheng He and Yuxuan Wang and Xian Shi and Ting He and others},
  journal={arXiv preprint arXiv:2509.17765},
  year={2025}
}

@string{acmmm = "Proc. ACM MM"}

@string{cvpr   = "Proc. CVPR"}

@string{icassp="Proc. ICASSP"}

@string{iclr = "Proc. ICLR"}

@string{icml =  "Proc. ICML"}

@string{is =  "Proc. Interspeech"}

@string{naacl = "Proc. NAACL"}

@string{neurips = nips}

@string{acl = "Proc. ACL"}

@string{emnlp = "Proc. EMNLP"}

@string{tpami =  "IEEE Trans. on Pattern Analysis and Machine Intelligence"}

@string{cvprw   = "Proc. CVPR Workshop"}

@string{aclf = "Findings of ACL"}
\endgroup


\end{document}